\documentclass{article}
\usepackage{spconf,amsmath,graphicx}
\usepackage{xcolor}
\usepackage{float}

\title{3M3D: Multi-view, Multi-path, Multi-representation for 3D Object Detection}
\name{Jongwoo Park\textsuperscript{ 1,2*}, Apoorv Singh\textsuperscript{ 1*}, Varun Bankiti\textsuperscript{ 1} \thanks{*Equal Contribution. This work was done while Jongwoo was interning at Motional.}}
\address{\textsuperscript{1}Motional,  \textsuperscript{2}Stony Brook University}
\begin{document}
%
\maketitle
\begin{abstract}
3D visual perception tasks based on multi-camera images are essential for autonomous driving systems. The latest work in this field performs 3D object detection by leveraging multi-view images as an input and iteratively enhancing object queries (object proposals) by cross-attending multi-view features. However, individual backbone features are not updated with multi-view features, and it stays as a mere collection of the output of the single-image backbone network. Therefore we propose \emph{3M3D: A Multi-view, Multi-path, Multi-representation for 3D Object Detection} where we update both multi-view features and query features to enhance the representation of the scene in both fine panoramic view and coarse global view. Firstly, we update multi-view features by multi-view axis self-attention. It will incorporate panoramic information in the multi-view features and enhance understanding of the global scene. Secondly, we update multi-view features by self-attention of the Region of Interest (ROI) windows which encodes local finer details in the features. It will help exchange the information not only along the multi-view axis but also along the other spatial dimension. Lastly, we leverage the fact of the multi-representation of queries (MRQ) in different domains to further boost performance. Here we use sparse floating queries along with dense Bird's Eye View (BEV) queries, which are later post-processed to filter duplicate detections. Moreover, we show performance improvements on the nuScenes benchmark dataset \cite{nuscenes} on top of our baselines. 
\end{abstract}
\section{Introduction}
\label{sec:intro}
3D object detection from visual information is a long-standing challenge for low-cost 2D sensors, i.e., cameras. While object detection from point clouds collected using more-expensive sensors like LiDAR benefits from information about the 3D structure of the visible objects, the camera-based setting has ill-posed structural information. With cameras, we must generate 3D bounding box predictions solely from the 2D information contained in the 2D RGB images.

\begin{figure}[htb]
  \centering
  \centerline{\includegraphics[width=8.5cm]{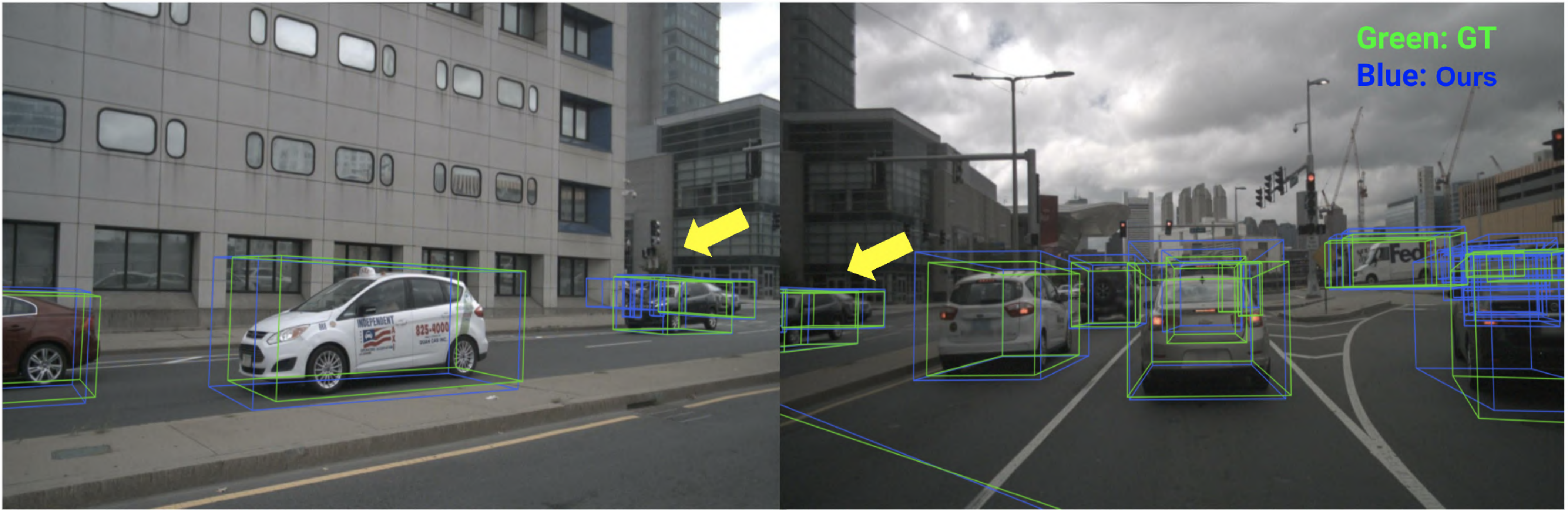}}
\caption{A multi-camera snippet from nuScenes \cite{nuscenes} showing \emph{3M3D} can detect objects with the tight bounding boxes, even if the objects are partially visible in a particular camera (Highlighted with yellow camera).GT: Ground-truth data.}
\label{fig:vis_1}
\end{figure}

\begin{figure*}[htb]
  \centering
  \centerline{\includegraphics[width=17cm]{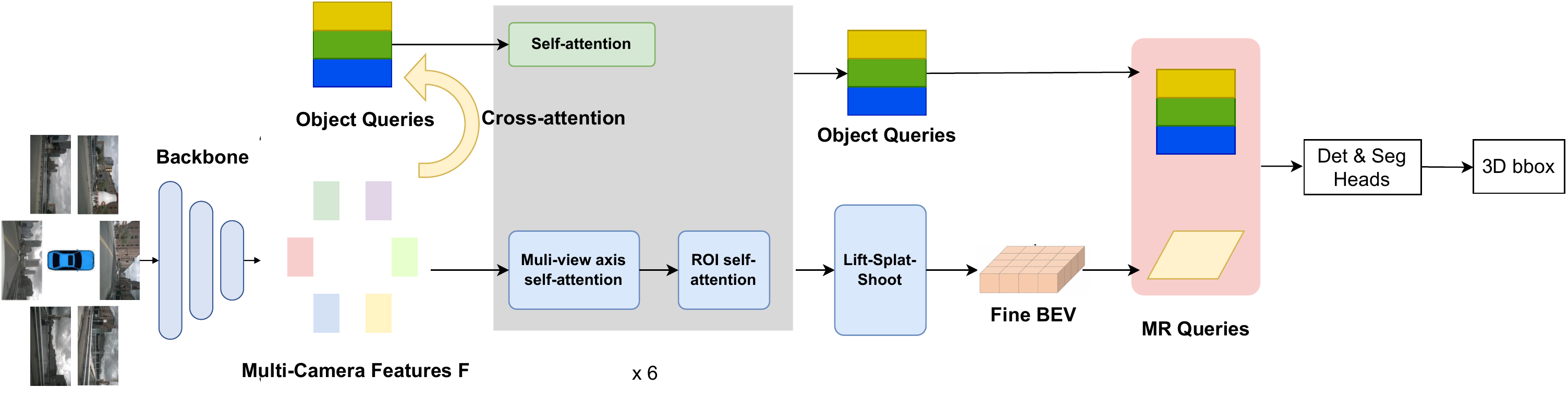}}
\caption{Overall architecture of \emph{3M3D: Multi-view, Multi-path, Multi-representation}. (Left-to-right) A) Multi-view camera images from nuScenes\cite{nuscenes} data. B) Individual image is passed through 2D backbone, and multi-camera features \emph{F} are collected. C) \emph{F} features pass through self-attention layers as defined in section \ref{sec:self-attn-1} and \ref{sec:self-attn-2} in the gray color-coded block. This block is sequentially stacked six times. D) Sparse floating queries \cite{detr3d} and dense fixed queries \cite{bevformer} are generated and aggregated together as defined in section \ref{sec:multi_query}. It is represented in \emph{MR: Multi-representation queries} in the yellow color-coded block. E) Finally, 3D bounding boxes are predicted after NMS decoding to filter duplicate detections.}
\label{fig:architecture}
\end{figure*}

Most straight forward way to perform 3D object detection is by using a monocular image-based detection paradigm \cite{detr}\cite{fcos3d}\cite{ssd}, consequently performing the post-processing step outside the ML model to aggregate information from all multi-camera system. However, this paradigm suffers from the limited global information availability in the backbone and the detection head. \\
A more unified framework has also been explored, where individual backbone features are extracted from the multi-views backbones separately, and subsequently, this information is fused together in the detection head \cite{detr3d}\cite{bevformer}\cite{petr}. As we established before, camera-based BEV (Bird's Eye View) detectors rely on learned priors and not the 3D structural information from sensors - just the detector's head-level feature aggregation limits the network to encode enough global information before making holistic-scene-aware predictions. \\ \\
Our work addresses these gaps with the following contributions:
\begin{itemize}
\item We introduced \emph{Multi-view self-attention layer}, which encodes global (panoramic) information inside the backbone features under the assumption that different views at the same height should have highly correlated information about the autonomous vehicle surroundings. Thereby creating panorama-aware features.
\item We also introduced \emph{ROI (Region of interest) self-attention layer} that encodes nearby pixels information by self-attending to each other using learned key pixel locations to save computation cost.
\item Lastly, we introduced \emph{Multi-representation queries},  which aggregates two different sets of queries viz., sparse floating queries\cite{detr3d} and dense fixed grid queries \cite{lss}\cite{bevformer} together to leverage their individual benefits.
\end{itemize}

\section{Related Work}
\subsection{Single-view Based Methods}
RCNN \cite{rcnn} pioneered the work for object detection using deep learning. It follows a two-stage object detection paradigm, where the first-stage object proposals are predicted, and later refined in the second stage. The high computation cost of this paper was later addressed in \cite{fasterrcnn}. In the subsequent single-stage line of work, object detection was explored with SSD \cite{ssd} using anchor-boxes heuristics. Later these heuristics were let go using object-center-based object detection paradigm in \cite{fcos3d}. \cite{detr} \cite{sv_survey} \cite{vision_radar} introduced transformer framework from language models in computer vision task. These methods are at-par or even better in some cases compared to traditional CNN-based approaches for object detection tasks. 

Even though motivated by temporal patches, TimeSformer \cite{timesformer} introduced a smart query-key pairs strategy to reduce the complexity of the problem. They align image patches temporally and spatially to limit the attention and compute resources while improving on performance accuracy. 
\subsection{Multi-view Based Methods}
DETR3D \cite{detr3d} extends the idea of DETR \cite{detr} approach to multi-view images and performs 3D object detection. It used a top-down approach where 3D position extracted from queries is used to sample 2D features from the backbone. It used camera transformation matrices to back-project 3D points onto the camera 2D feature map. Authors claim that sparse representation of the queries enables us to skip the post-processing step of Non-maximum Suppression (NMS). BEVFusion\cite{bevfusion} and BEVFormer \cite{bevformer} follow a similar approach; however, they define 3D queries in a more structured format on a BEV grid. Here each cell in the grid is treated as an object proposal. It also encodes temporal information via a self-attention layer attending to temporal BEV features. \\ PETR \cite{petr} encodes 3D coordinate information in the 2D backbone features itself, to get 3D position-aware features. The object queries directly interact with 3D position-aware features and output 3D detection results here. LSS: Lift, Splat, Shoot\cite{lss} approaches this problem a little differently, where they first generate a pseudo-point-cloud from the pixels by associating discretized depth for each pixel. On top of this point cloud, any standard point-cloud-based 3D detection head can be used.

\section{Method}
In this work, we introduced three novel transformer-based layers on top of the baseline DETR3D \cite{detr3d} work, as shown in Fig. \ref{fig:architecture}. Let us go through each of them in detail: 

\begin{figure*}[htb]
  \centering
  \centerline{\includegraphics[width=\linewidth]{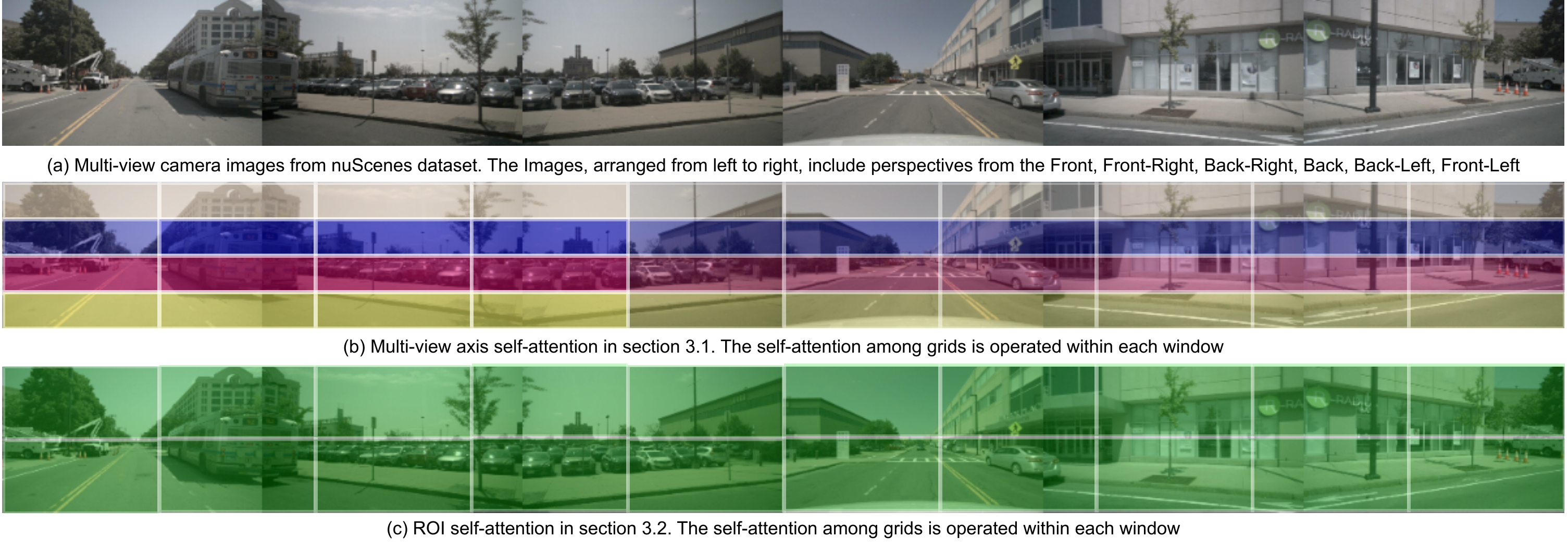}}
\caption{Multi-view axis and ROI self-attention on the panoramic image. In multi-view axis self-attention, windows are shifted horizontally to exchange information among features at the same height. In ROI self-attention, windows are shifted diagonally to exchange information among nearby features.}
\label{fig:self-attn}
\end{figure*}

\subsection{Multi-View Axis Self-Attention Layer}
\label{sec:self-attn-1}
Inspired by TimeSformer\cite{timesformer}, but applying the concept on multi-views instead of time-series data, we break down pixels-patch attention region for more targeted information sharing and lowering complexity burden. In addition to Timesformer\cite{timesformer}, CSWin\cite{cswin_Dong_2022_CVPR} also shows that self-attention between grids at the same height but different widths works well. This is visualized in row-2 of Fig. \ref{fig:self-attn}, where we apply a self-attention layer on color-coded areas. For example, self-attention in each color only focuses on features present in multi-views at a specific height range. This assumption is based on the fact that all the cameras (panorama) will have some correlation along the height dimension of their images, hence exchanging information across camera views only in a particular height range. Further breaking down the entire height patch on panoramic windows using small windows, as shown with adjacent same color-coded blocks in row-2 Fig. \ref{fig:self-attn} reduced complexity burden. With every iterative self-attention layer (Total 6), these color-coded blocks are shifted horizontally by a bit to further encode information at a greater distance. In the experiment, the multi-scale feature maps are divided into windows of different sizes: (3, 32), (3, 32), (3, 32), and (3, 24), arranged in descending order based on the feature map resolution. These windows represent the height and width values of each respective window. The computation complexity of the window self-attention is $$\frac{1}{r}\cdot \mathcal{O}(B\cdot(MHW)^2\cdot C)$$ where r is the number of windows created in a feature map, B is the batch-size, M is the number of cameras, H and W are the width and height of the feature map, C is the channel of the latent vector. Our method is only 0.2\% of the full self-attention. 

\subsection{ROI (Region of Interest) Self-Attention Layer}
\label{sec:self-attn-2}
To further refine backbone features with local information, we perform another self-attention layer to a spatial ROI within a camera view. The motivation to do this was that pixel information within a spatial patch, viz., ROI, will encode local information very well and add context to immediate neighboring pixels for large objects. Swin\cite{swin_Liu_2021_ICCV} demonstrates that objects in an image can be extracted by self-attending grids within a window as they are spatially close to each other. Ten classes (car, truck, bus, trailer, construction vehicle, pedestrian, motorcycle, bicycle, cone, barrier) in the Nuscene dataset are more or less like a square shape, so a square window is suitable to formulate a rich representation of those objects. ROI windows are shown with color-coded blocks in row-3 of Fig. \ref{fig:self-attn}. Here we also leverage the concept of shifted windows\cite{timesformer} with every six sequential attention layers to further expand the receptive field of the information exchange. In the experiment, multi-scales feature maps are divided into windows of different sizes: (12, 12), (12, 12), (6, 6), and (9, 12), arranged in descending order based on the feature map resolution. For example, at the feature map (72, 768) with the highest resolution, 384 windows (regions) are created and self-attention is performed within each window. At every alternate stage, windows are shifted by (6, 6), (6, 6), (3, 3), and (0, 0), respectively. The computation complexity of our method is only 0.2\% of the full self-attention. Results of this, along with the self-attention layer from section \ref{sec:self-attn-1}, are shown in table \ref{table:result_1}.

\begin{table*}[t]
\caption{\small self-attention layer as defined in section \ref{sec:self-attn-1} and \ref{sec:self-attn-2} compared to DETR3D baseline \cite{detr3d}. These results are evaluated on nuScenes test set. The unit of relative improvement in the parentheses is \%, and the metric key is defined in section \ref{sec:results}. Trained on 48 epochs.}
\vspace{-2.0mm}
\begin{center}
\resizebox{2.0\columnwidth}{!}{
\begin{tabular}{|l|r|r|r|r|r|r|r|r|}
\hline
Methods & NDS(\%) $\uparrow$ & mAP(\%) $\uparrow$ & mATE(cm) $\downarrow$ & mASE(\%) $\downarrow$ & mAOE(rad) $\downarrow$ & mAVE(cm/s) $\downarrow$ & mAAE(\%) $\downarrow$ \\
\hline
DETR3D \cite{detr3d} & 40.1 & 31.4 & 77.9 & 27.0 & 0.44 & \textbf{88.2} & \textbf{19.1} \\
3M3D (Ours) & \textbf{(+2.1) 41.0} & \textbf{(+4.3) 32.7} & \textbf{(+0.1) 77.8}& \textbf{(+0.3) 26.9} & \textbf{(+0.8) 0.40} & (-1.6) 89.6 & (-3.7) 19.8 \\
\hline
\end{tabular}
}
\label{table:result_1}
\end{center}
\end{table*}

\begin{table*}[t]
\vspace{-5.0mm}
\caption{\small Average precision per class, NDS, and mAP of multi-representation of queries as defined in section \ref{sec:multi_query} compared to DETR3D baseline \cite{detr3d} are presented. These results are evaluated on nuScenes test set. The unit of relative improvement in the parentheses is \%, and the metric key is defined in section \ref{sec:results}. Trained on 12 epochs.}
\vspace{-5.0mm}
\begin{center}
\resizebox{2.08\columnwidth}{!}{
\begin{tabular}{|l|r|r|r|r|r|r|r|r|r|r|r|r|r|}
\hline
Methods & NDS(\%) $\uparrow$  & mAP(\%) $\uparrow$& car(\%) $\uparrow$& truck(\%) $\uparrow$ & bus(\%) $\uparrow$ & trailer(\%) $\uparrow$ & conv.(\%) $\uparrow$ & pedes.(\%) $\uparrow$ & motoc.(\%) $\uparrow$ & bicycle(\%) $\uparrow$ & cone(\%) $\uparrow$ & barrier(\%) $\uparrow$
\\
\hline
DETR3D \cite{detr3d} & 38.4 & 31.2 & 51.2 & 25.8 & 34.6 & 13.1 & 6.4 & 37.9 & 29.3 & 22.8 & 47.2 & \textbf{44.7} \\
3M3D (Ours) & \textbf{(+0.9) 38.7} & \textbf{(+2.7) 32.0} & \textbf{(+1.6) 52.0} & \textbf{(+3.9) 26.8} & (+0.0) 34.6& \textbf{(+16.8) 15.3} & (+0.0) 6.4 & \textbf{(+0.5) 38.1} & \textbf{(+2.0) 29.9} & \textbf{(+10.5) 25.2} & \textbf{(+0.2) 47.3} & (-0.2) 44.6 
\\
\hline
\end{tabular}
}
\label{table:result_2}
\end{center}
\end{table*}


\subsection{Multi-representation Queries}
\label{sec:multi_query}
For the final novel layer, inspired by work M3DETR\cite{m3detr} where they fuse liDAR point-cloud input but in different representations viz., point, voxels, and features and see a boost in performance, we applied this multi-representation framework to our queries. Our first representation is sparse floating queries that we get natively from \cite{detr3d} baseline work. These queries are sparse queries that are representative of the entire training data. Our other representation is inspired by \cite{bevformer} and \cite{lss}, which use densely fixed queries on the BEV grid; these are natively the size of $128^2$. We selected top-500 fixed BEV queries based on the confidence level of the original $128^2$ BEV queries. We use these top-500 fixed and 900 floating queries to conduct experiments in table \ref{table:result_2}. 

\section{Experiments and Results}
\subsection{Experiment Settings}
Following previous methods, we utilized ResNet101-DCN\cite{resnet} backbone initialized from FCOS3D\cite{fcos3d} checkpoint. By default, we utilized the output from multi-scale features from FPN (Feature Pyramid Network)\cite{fpn} with sizes of $1/16$, $1/32$, $1/64$ of original image resolution and the dimension of $C=256$. Each training job job ran on eight parallel Nvidia V100 GPUs, with 48 and 12 epochs in table \ref{table:result_1} and \ref{table:result_2}, respectively. Input image resolution per camera was $(576, 1024)$. The NMS threshold set for filtering duplicate detections is $0.2$. In section \ref{sec:multi_query}, DETR3D\cite{detr3d} like head used 900 queries, and BEVFusion\cite{bevfusion} like head used $128^2$ queries, out of which 500 top-k queries, based on confidence values of predictions, was also experimented in table \ref{table:result_2}.

\subsection{Results}
\label{sec:results}
We evaluated our method on nuScenes \cite{nuscenes} benchmark dataset, which has six multi-view calibrated cameras. Train, validation, and test set includes 28k, 6k, and 6k samples, respectively, with 1.4M annotated 3D bounding boxes. For evaluation metric comparison, we use mean-average precision (mAP) and nuScenes Detection Score (NDS), calculated with center-distance matching criteria between predictions and ground truths. In addition, we also share results on True-positive (TP) metrics defined in the dataset viz., mATE: mean Average Translation Error; mASE: mean Average Scale Error; mAOE: mean Average Orientation Error; mAVE: mean Average Velocity Error; mAAE: mean Average Attribute Error. Self-attention layers in section \ref{sec:self-attn-1} and \ref{sec:self-attn-2} have shown $4.3\%$ improvement on mAP and $8.0\%$ improvement on mAOE, showing overall 3D object detection along with orientation improved with the better collaboration of feature pixels in self-attention layer as shown in table \ref{table:result_1}. Multi-representation queries in section \ref{sec:multi_query} resolve class-imbalance problem as shown in table \ref{table:result_2}. It improves the average precision of rare classes such as trailers, bicycles, trucks, motorcycles, and cones by 16.8\%, 10.5\%, 3.9\%, 2.0\%, and 0.2\%. Overall, it improves mAP by 2.7\% and NDS by 0.9\%. We believe that BEV queries are assigned to the same location in every iteration so that rare objects are well-captured when they are targeted by BEV queries. The throughput of the original DETR3D baseline \cite{detr3d} is 35 imgs/s and our model is 31 imgs/s. Therefore, our model improves NDS, mAP by +0.9\%, +2.7\%, respectively, while 12\% slower throughput. The setting to measure the throughput is: 
Number of GPU = one NVIDIA A10 24GB, Batch size=2, Samples per iteration=50. It is useful for the severely class-imbalanced dataset. For example, in the Nuscene dataset, six rare classes compose less than 50\% of the total instances. In contrast, we confirmed no significant improvement in rare classes and a -0.5\% 
\vspace{-0.4cm}
\\
\section{Conclusion}
In this work, we present an architecture for enhancing backbone features with a multi-view axis and ROI self-attention layers, which iteratively refines them with added context. Our model outperforms our baseline \cite{detr3d} significantly, as shown in table \ref{table:result_1} and \ref{table:result_2}. We also visualize our self-attention windows for an easier understanding of the framework in Fig. \ref{fig:self-attn}. We also show aggregation of different query representations to improve further the performance of our baseline network in section \ref{sec:multi_query}.

\bibliographystyle{IEEEbib}
\bibliography{refs}

\end{document}